# Knowledge Graph-Guided Temporal Behavior Interpretation with a Local LLM Agent in Smart Homes

Tianle Hu
*Department of Information, Artificial Intelligence and Data Science*
*Daiichi Institute of Technology*
Tokyo, Japan
0009-0007-5929-6546

Hong Chen
*School of Artificial Intelligence*
*Xiamen Institute of Technology*
Xiamen, China
0000-0001-7540-3021

Yenjou Wang
*Department of Information, Artificial Intelligence and Data Science*
*Daiichi Institute of Technology*
Tokyo, Japan
0000-0002-1319-2190

***Abstract*—With the rise in aging populations and single-person households, privacy-preserving smart-home behavior monitoring is increasingly essential for daily living analysis and home-care applications. Smart-home sensors continuously record daily activities; however, these data are typically stored as low-level event logs, which are not directly interpretable as meaningful behavioral patterns. This paper presents a knowledge graph (KG)-guided local large language model (LLM) agent for temporal behavior interpretation in smart homes. The proposed system utilizes Home Assistant sensor histories and Qwen2.5-0.5B-Instruct, a lightweight local LLM, to interpret and explain daily behavior patterns. Sensor records are transformed into behavior-level summaries and organized into rolling-window features to capture multi-day changes. The proposed framework further adapts Qwen2.5 using Low-Rank Adaptation (LoRA) with KG-guided input. Experimental results using an external dataset and local Home Assistant data demonstrate that KG-guided fine-tuning improves the stability and consistency of temporal-state interpretation compared to the base model. Additionally, five-day and seven-day windows yield more stable KG-guided interpretations than shorter windows.**

***Keywords*—Smart Home, Home Assistant, Large Language Model, Knowledge Graph, Temporal Behavior Interpretation, Rolling Window**

## I. Introduction

The increasing prevalence of aging populations, single-person households, and the demand for home-care services has intensified the need for technologies capable of monitoring daily living patterns in residential environments. Smart-home sensing systems address this need by collecting activity records from presence sensors, motion sensors, contact sensors, and other IoT devices. As illustrated in Fig. 1, these sensors generate low-level event records that can be organized into daily behavior summaries for higher-level analysis. Given that these records reflect private daily routines, their interpretation must prioritize user privacy. Several challenges persist in interpreting smart-home behavior data. First, most smart-home data are stored as low-level event histories or device states, which are difficult to interpret directly as meaningful changes in a resident's living pattern [1]. Second, isolated sensor events or single-day summaries are insufficient for understanding changes in temporal behavior [2]. For instance, a decrease in activity on a single day may reflect a temporary schedule change, whereas a similar pattern over several consecutive days may indicate a more stable behavioral shift. Third, although local lightweight large language models (LLMs) are promising for privacy-aware processing, their limited model capacity and reliance on numerical inputs may result in unstable or weakly grounded interpretations.

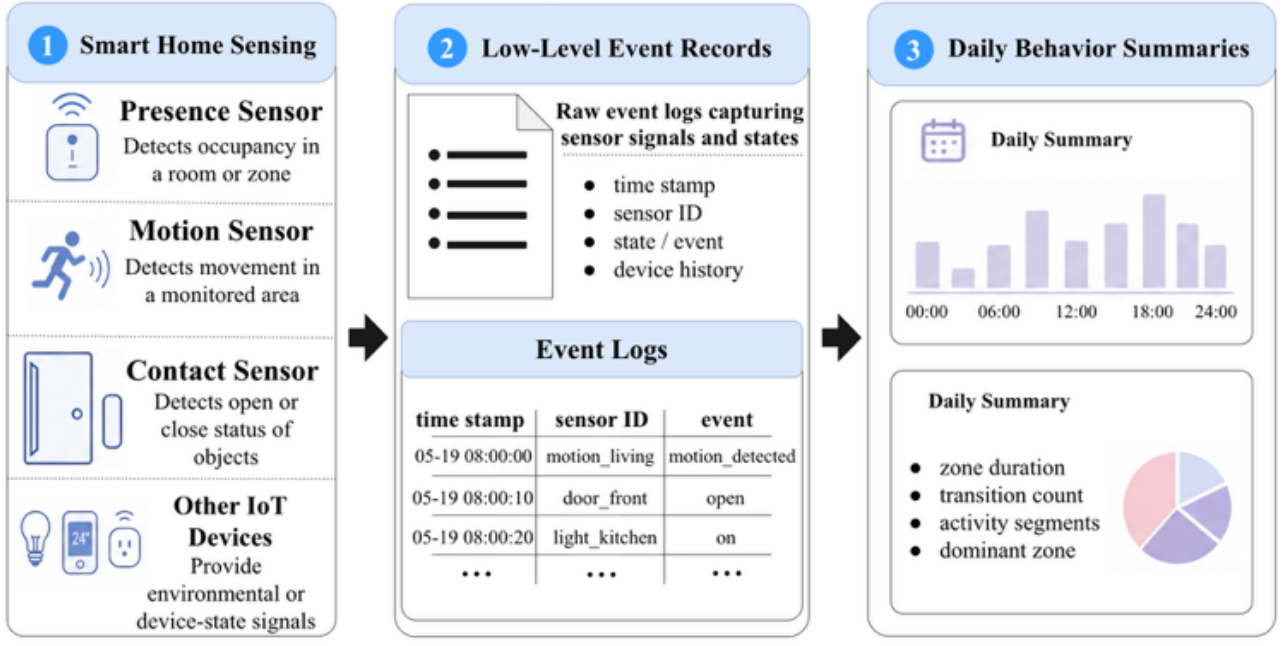

Fig. 1. Conceptual flow from smart-home sensing to daily behavior summaries

Without explicit temporal and contextual grounding, the model may focus on local feature changes and overlook broader multi-day behavior patterns. Consequently, this study addresses the problem of interpreting multi-day smart-home behavior changes in a local, privacy-aware, and context-grounded manner.

To address these challenges, this paper proposes a knowledge graph (KG)-guided, lightweight, local LLM framework for interpreting temporal behavior in smart homes. The method converts sensor histories into daily behavior summaries and groups consecutive summaries into rolling-window samples. A supervised fine-tuning dataset is constructed to adapt Qwen using Low-Rank Adaptation (LoRA) [3], yielding the P95-LoRA Qwen model. The main contributions of this study are summarized as follows. First, we develop a privacy-aware, local LLM framework for interpreting smart-home behavior with fully local processing. Second, we combine rolling-window temporal representations with KG-based evidence to capture multi-day behavior changes and support stable interpretation. Third, the proposed framework is evaluated using both the CASAS Aruba dataset [4] and locally collected Home Assistant data on a Raspberry Pi, demonstrating improved consistency and stability compared with the base Qwen model.

## II. Related Work

### A. Smart-Home Sensing and Temporal Behavior Monitoring

Smart-home sensing has been widely studied for observing activities of daily living in residential environments. Multi-sensor residential datasets have shown that indoor sensing systems can capture activity, location, posture, and movement-related behavior in home-like settings [1]. The CASAS smart-home project has also provided long-term ambient sensor event streams widely used for activity recognition and behavior analysis in residential environments [4]. Recent studies have further considered layout differences, temporal context, behavioral noise, and zero-shot recognition in realistic smart-home environments [5]–[7]. However, most existing studies mainly focus on activity recognition, anomaly detection, or dataset construction. This limitation motivates the need for temporal behavior interpretation that explicitly considers changes over time.

### B. LLM and KG-Guided Interpretation for Smart Homes

Another important issue in smart-home behavior analysis is how to make sensor-derived results understandable and reliable for users. LLMs have recently been explored as interfaces for producing natural-language explanations and activity interpretations from sensor-based evidence. Recent studies have explored LLM-based explanation and zero-shot activity interpretation from smart-home sensor data [8]–[10]. Nevertheless, LLM-based interpretation requires structured grounding when applied to smart-home sensor data. Without explicit context, a model may focus on a local feature change and generate unstable or weakly supported explanations. KGs provide a structured way to represent entities, relations, and contextual information [11]. Ontology- and KG-based approaches have also been used to support smart-home reasoning and contextual grounding for language models [12]–[14]. Existing LLM-based and KG-based smart-home studies mainly focus on activity recognition, explanation generation, device reasoning, or context-aware activity modeling. They often use structured knowledge to represent activity categories, device relations, environmental context, or common-sense activity constraints. Despite these advances, the representation and interpretation of temporal behavior changes across consecutive days remain underexplored, especially when the input is given as numerical daily summaries for a lightweight local model. Therefore, smart-home behavior interpretation requires a framework that combines local LLM inference with structured temporal evidence.

## III. Proposed Method

This section presents the proposed pipeline for transforming low-level smart-home sensor histories into interpretable temporal behavior descriptions. Section III-A describes the overall local interpretation framework, Section III-B introduces the rolling-window representation for capturing multi-day behavior changes, and Section III-C explains the KG-guided local LLM interpretation method for guiding the lightweight model with structured temporal evidence.

### A. Overview of the Framework

This section introduces the overall structure of the proposed KG-guided local LLM framework, which aims to interpret multi-day behavioral changes from smart-home sensor histories. As shown in Fig. 2, the framework comprises five main stages. First, Home Assistant collects sensor histories and maps the monitored areas into semantic zones. The collected histories are then converted into daily behavior summaries containing zone-level and activity-related features. Subsequently, consecutive daily summaries are grouped into rolling windows to capture temporal changes across multiple days. Based on the aggregated features and baseline comparisons, the system generates structured temporal states and organizes them into a lightweight KG context. Finally, the local LLM employs the daily summaries, rolling-window features, and KG context to predict the temporal state and generate a natural-language explanation.

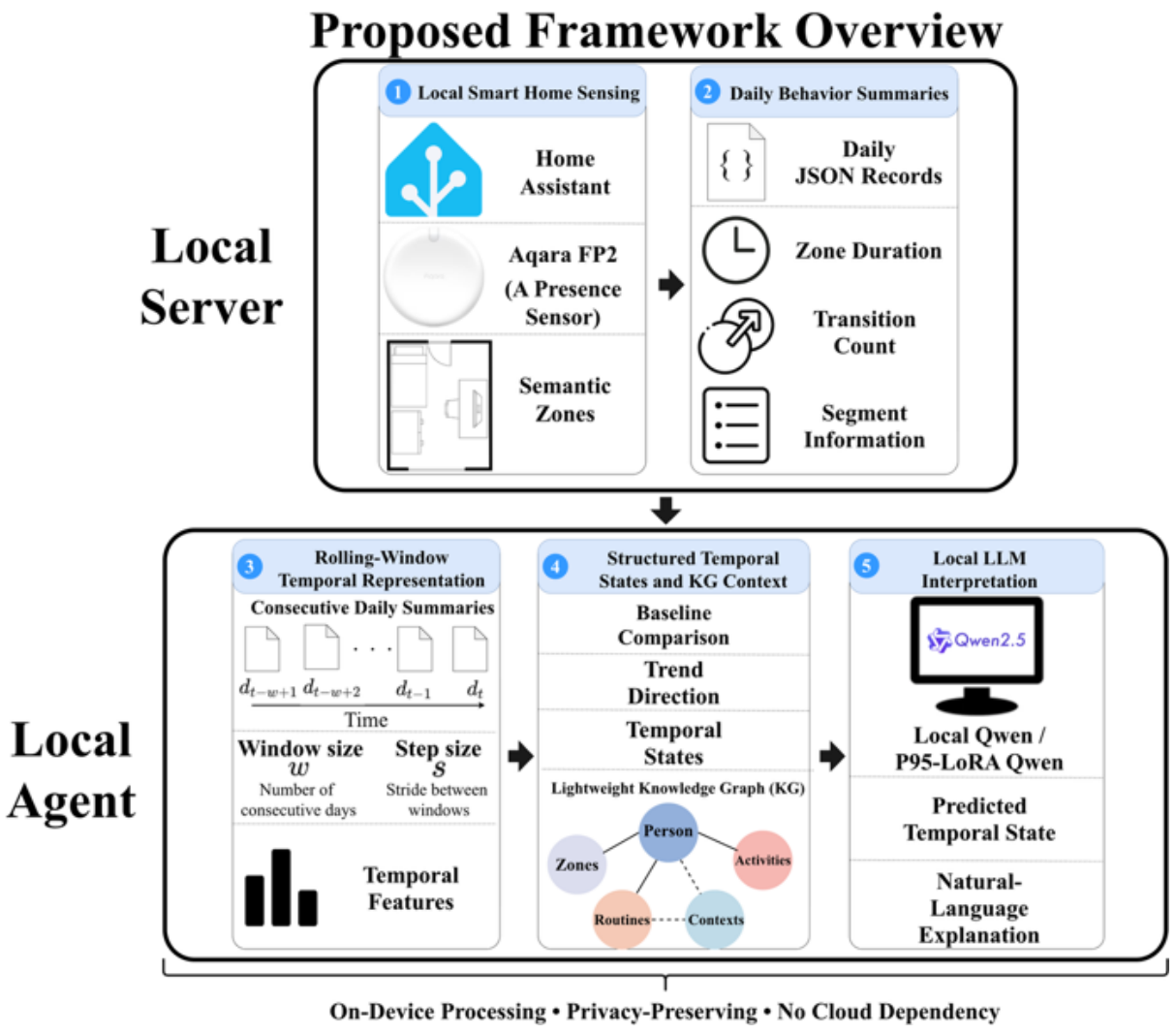

Fig. 2. Overview of the proposed KG-guided local LLM framework for temporal behavior interpretation in smart homes.

### B. Rolling-Window Temporal Representation

A rolling-window representation transforms consecutive daily behavior summaries into temporal samples. $X = \{x_1, x_2, \dots, x_n\}$ denote a sequence of $n$ daily behavior summaries, where $x_d$ represents the behavior record of day $d$. Each daily summary contains behavior-level features such as zone-level duration, movement transitions, and segment-level activity information.

For a window index $t$, the rolling window $W_t$ is defined as

$$W_t = \{x_{(1+(t-1)s)}, x_{(2+(t-1)s)}, \dots, x_{(w+(t-1)s)}\} \tag{1}$$

The total number of rolling windows is calculated as

$$k = \left\lfloor \frac{n-w}{s} \right\rfloor + 1 \tag{2}$$

In this formulation, the window size $w$ determines how many consecutive days are included in one temporal sample, while the step size $s$ determines how far the window moves between adjacent samples. The floor operation discards any incomplete window remaining after the final shift, and the additional $+1$ includes the initial window starting from the first day. A smaller window can capture short-term fluctuations, whereas a larger window provides more stable temporal context. Rolling-window strategies have also been used in recent time-series studies to construct fixed-length temporal samples from sequential data [15].

For each rolling window $W_t$, temporal features are extracted by aggregating the daily summaries within the window. These features include mean zone duration, transition frequency, dominant zone tendency, and changes in activity segments. The framework also compares the current window with a recent behavioral baseline to determine whether the behavior remains close to the usual pattern or deviates from it.

Based on these temporal features, the framework generates structured temporal states, such as close to baseline, increasing trend, decreasing trend, fragmented behavior, and short-term fluctuation. These temporal states are not manually annotated labels, but are derived from observable sensor-based behavior features. They are later used as structured evidence for KG-guided LLM interpretation.

### C. KG-Guided Local LLM Interpretation

After structured temporal states are generated from rolling-window samples, they are organized into a lightweight KG context. This KG is not used as a large-scale general ontology. Instead, it serves as a compact representation of temporal behavior evidence, including zone-level behavior, trend direction, baseline comparison, and the generated temporal state.

For each rolling window $W_t$, the KG context $G_t$ is constructed from the extracted temporal features and temporal states. It can be represented as

$$G_t = (V_t, E_t) \quad (3)$$

where $V_t$ denotes the set of nodes and temporal states, and $E_t$ denotes the relations among them.

Fig. 3 illustrates the schema of the lightweight temporal KG, including the main entity types and their relations. This structure allows the system to provide explicit evidence to the LLM instead of relying only on numerical features or raw sensor summaries.

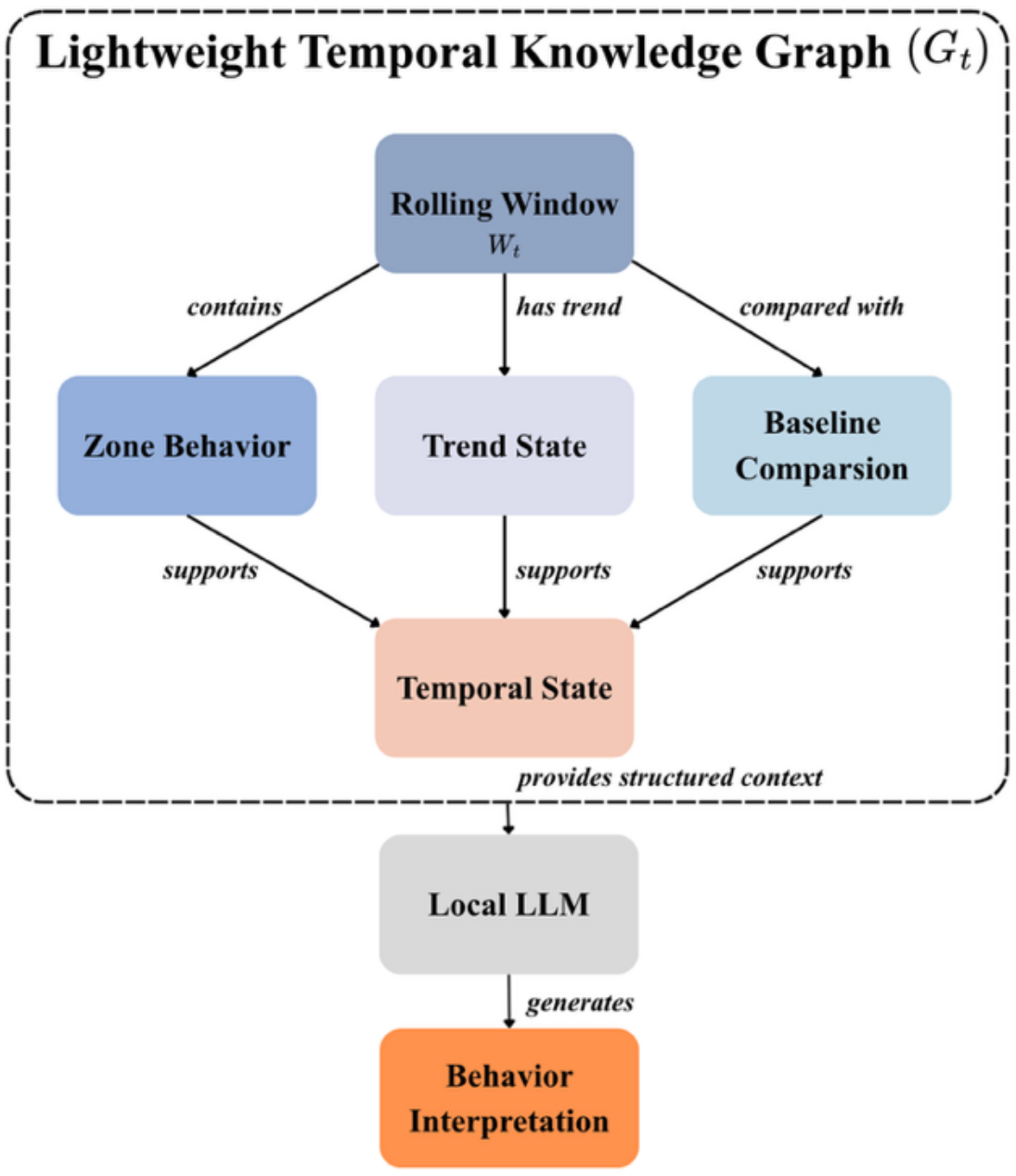

Fig. 3. Schematic of the lightweight temporal KG and its main entity–relation structure.

The structured temporal evidence provides grounding for the interpretation, while the local LLM primarily synthesizes the heterogeneous evidence into a coherent natural-language explanation of recent behavior changes. The predicted temporal state is used as an intermediate indicator of whether the generated interpretation remains aligned with the structured evidence. This process can be described as

$$(\hat{y}_t, r_t) = LLM(W_t, G_t) \quad (4)$$

where $\hat{y}_t$ is the predicted temporal state and $r_t$ is the generated explanation. By using KG-based context, the framework aims to make the LLM interpretation more consistent with sensor-derived temporal evidence.

## IV. Experiment

### A. Local Smart-Home Environment

The experimental environment was built on a local Home Assistant-based smart-home setup. Home Assistant was deployed on a Raspberry Pi 400 device equipped with a quad-core ARM Cortex-A72 CPU and 4 GB RAM, and used as the central platform for collecting and managing sensor states. An Aqara FP2 presence sensor was integrated into Home Assistant to capture room-level presence information. The Aqara FP2 sensor areas were mapped to semantic zones, such as the rest area, entry, storage area, work area, and window side, as shown in Fig. 4. These zones served as the spatial units for converting raw presence-sensor histories into daily behavior summaries.

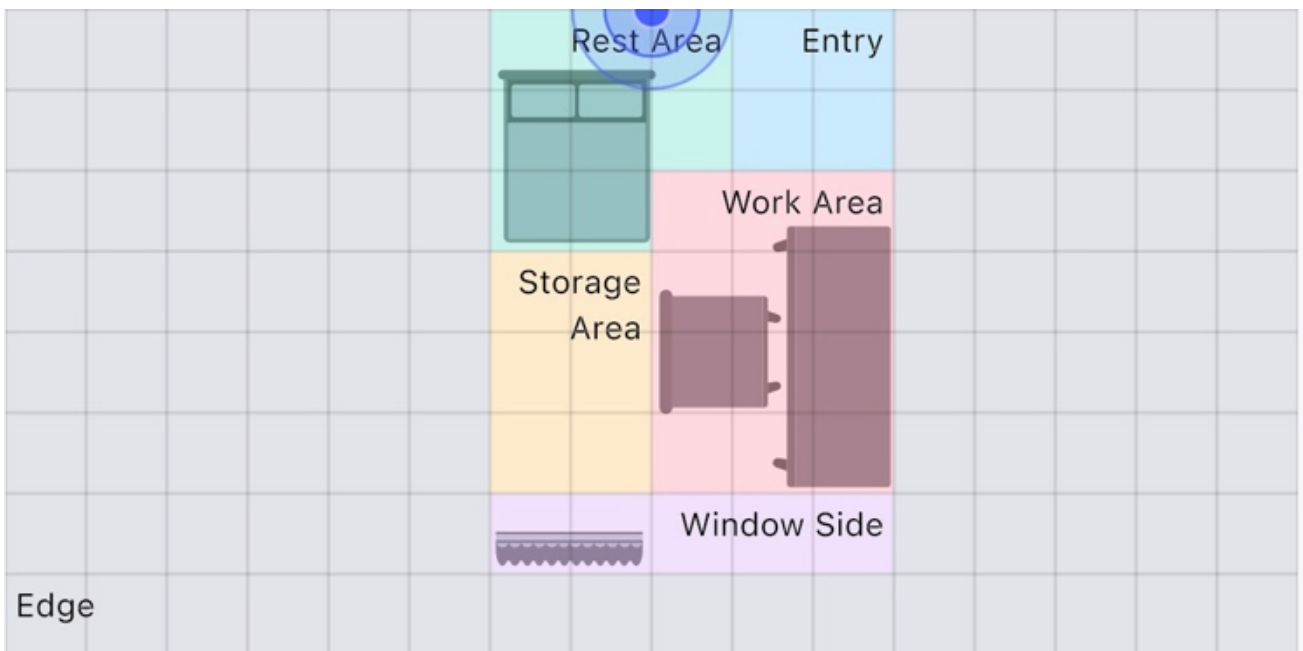

Fig. 4. Local smart-home layout and semantic zones.

To support local behavior interpretation, we implemented a local agent service as a Home Assistant add-on. The service monitored Home Assistant state histories, generated daily behavior summaries, and stored rolling-window behavior records for later analysis. In the local pipeline, the collected Home Assistant state histories were converted into daily JSON records. This format retains behavior-level information while avoiding direct dependence on raw event logs during model inference. These records were stored locally and used as the input for subsequent temporal analysis.

### B. External Dataset and Data Preprocessing

To examine whether the proposed rolling-window temporal interpretation method can be applied beyond our local Home Assistant prototype, we used the Aruba dataset from the CASAS smart-home project as an external dataset. Aruba records the daily life of a single resident in a sensor-equipped home from November 2010 to June 2011, covering approximately 220 days of continuous observations. Since our study focuses on temporal changes in individual living patterns, this dataset is suitable because it provides long-term sensor records from a solo living environment. The raw Aruba data consist of time-stamped sensor events generated mainly from motion sensors and contact sensors. In our preprocessing, these low-level events were converted into daily behavior summaries. Sensor events were grouped into semantic zones, and daily features such as zone duration, transition count, and activity segments were extracted.

### C. Percentile-Based Threshold Selection

Before constructing rolling-window samples, percentile-based filtering was applied to the daily behavior summaries converted from the Aruba dataset. This preprocessing step reduced the influence of extreme records, such as unusually large total activity duration or abnormal zone-level duration. At the same time, the filtering process needed to preserve sufficiently diverse and difficult behavior patterns, because removing too many irregular records could reduce the generality of the subsequent interpretation model. We examined three single-threshold settings, Percentile 90% (P90), Percentile 95% (P95), and Percentile 99% (P99), where records exceeding the selected percentile threshold for major duration-based features were treated as anomalous. In addition, a combined Percentile 95% + Percentile 99% (P95 + P99) rule was tested to remove direct extreme outliers while also detecting records with multiple moderately abnormal features. Table I summarizes the effects of these filtering settings on clean daily records, rolling-window samples, and model performance. Among the four settings, P90 removed the largest number of daily records, leaving 158 clean days and 777 rolling-window samples. Although P90 achieved the highest LoRA test match rate of 69.70%, it also produced the lowest temporal state entropy, suggesting that strong filtering reduced the diversity of temporal behavior patterns. P95 achieved the second-highest LoRA test match rate of 67.78% while preserving more clean days and higher temporal state entropy than P90. In contrast, P99 preserved the largest number of clean days and rolling-window samples, but its LoRA test match rate was the lowest, at 55.61%. The combined P95 + P99 setting produced the largest number of KG-improved cases, indicating that this setting retained more cases where KG guidance affected the model interpretation. These results suggest that the filtering strategy influences both training data construction and model-level performance.

TABLE I. EFFECTS OF PERCENTILE-BASED ANOMALY FILTERING ON SFT CONSTRUCTION AND KG-GUIDED PERFORMANCE

| Filtering setting | Clean days | Anomalous days | Filtered ratio (%) | Rolling windows | Temporal state entropy | SFT rows | Base test (%) | LoRA test (%) | Improved by KG | Worsened by KG |
|---|---|---|---|---|---|---|---|---|---|---|
| P90 | 158 | 62 | 28.18 | 777 | 1.6555 | 612 | 0.00 | **69.70** | 0 | 0 |
| P95 | 185 | 35 | 15.91 | 912 | 2.0510 | 732 | 0.56 | 67.78 | 21 | 4 |
| P99 | 210 | 10 | 4.55 | 1037 | 2.3044 | 832 | 5.37 | 55.61 | 21 | 9 |
| P95 + P99 | 203 | 17 | 7.73 | 1002 | 2.2304 | 807 | 2.05 | 62.05 | **60** | 14 |

### D. Rolling-Window Construction and Window Size Analysis

Rolling-window analysis was used to capture temporal changes that a single-day summary cannot represent. Therefore, consecutive daily summaries were grouped into rolling windows, with each window serving as a temporal sample for feature extraction and interpretation. The window size determines the temporal scale of the analysis. A smaller window can respond quickly to short-term fluctuations, but it may also be sensitive to accidental daily variations. A larger window provides more stable temporal context, but it can smooth out rapid behavioral changes. For this reason, the window size was treated as an important parameter in the proposed temporal interpretation framework. To examine the effect of window size in a realistic setting, local Home Assistant data were continuously collected from April 6 to May 23, 2026, covering 48 days in total. The first 22 days were used for local model adaptation, and the subsequent 26 days were held out as an independent local test period for the window-size analysis reported in this study. Although the number of days was limited, the data reflected the actual target scenario of this study, namely daily behavior monitoring in a solo living space. Window sizes of 1, 2, 3, 5, and 7 days were compared. In particular, the five-day window approximately represents a weekday cycle, whereas the seven-day window represents a full weekly cycle. This comparison was used to examine whether a short, medium, or weekly-scale window was more suitable for KG-guided behavior interpretation.

TABLE II. WINDOW-SIZE COMPARISON USING LOCAL HOME ASSISTANT DATA

| Window Size | Baseline Parse | KG Parse | Baseline Evidence Consistency | KG Evidence Consistency | KG+ | KG− |
|---|---|---|---|---|---|---|
| 1 | 66.67% | 73.33% | 0.00% | 6.67% | 1 | 0 |
| 2 | 57.14% | 85.71% | 42.86% | 71.43% | 4 | 0 |
| 3 | 69.23% | 76.92% | 38.46% | 61.54% | 3 | 0 |
| 5 | 72.73% | **100.00%** | 27.27% | 81.82% | 6 | 0 |
| 7 | 88.89% | **100.00%** | 44.44% | 77.78% | 3 | 0 |

Table II summarizes the effects of window size on temporal interpretation. Parse indicates whether the model output could be converted into the required temporal-state format. Evidence Consistency measures whether the predicted temporal state is consistent with the structured reference, which is primarily grounded in human-provided contextual information and further refined using sensor-derived temporal evidence. KG+ and KG− denote the numbers of cases in which KG-guided prompting improved or worsened Evidence Consistency relative to the baseline prompt. The KG-guided prompt improved both parse success and Evidence Consistency across all window sizes. For the five-day window, KG Parse reached 100.00%, while KG Evidence Consistency increased from 27.27% to 81.82%, with six improved cases and no worsened cases. The seven-day window also achieved a KG Parse rate of 100.00% and a KG Evidence Consistency of 77.78%. These results indicate that five- and seven-day windows provide more stable temporal context than shorter windows while preserving interpretable behavioral changes. The structured temporal states generated from each rolling window were further organized as a lightweight KG context. This graph was constructed as a compact representation of temporal behavior relations rather than as a large-scale ontology. Each rolling window was associated with information such as dominant zone, duration trend, transition tendency, baseline deviation, and the corresponding temporal state. This KG-based context then guided the LLM prompt so the generated interpretation remained consistent with the sensor-derived temporal features.

### E. LoRA Fine-Tuning for Temporal Behavior Interpretation

The P95-filtered rolling-window samples and the KG-guided interpretation results described in the previous section were used to construct the supervised fine-tuning dataset. The P95 setting was selected because it preserved sufficient daily records and temporal-state diversity while reducing the influence of extreme sensor-derived records. Under this setting, 185 clean daily records were retained, and 912 rolling-window samples were generated. Each training instance was constructed from one rolling-window sample. The input contained the daily behavior summary, rolling-window temporal features, and KG-based temporal context. The target output was a natural-language interpretation of the resident's recent behavior pattern. In this way, the fine-tuning data connected sensor-derived temporal features with structured behavior explanations.

Qwen2.5-0.5B-Instruct was used as the base model [16]. Instead of full-parameter fine-tuning, LoRA [3] was applied to adapt the model to the temporal behavior interpretation task. LoRA adds a small number of trainable low-rank parameters while keeping most original parameters frozen. This makes the training process lightweight and suitable for a local smart-home agent setting. This paper refers to the fine-tuned model as P95-LoRA Qwen. Through LoRA fine-tuning, the model learned recurring relationships among daily behavior summaries, rolling-window temporal features, KG-based temporal states, and natural-language explanations. The trained P95-LoRA Qwen was then used to evaluate whether a lightweight LLM can learn the proposed temporal behavior interpretation framework. In the following evaluation, its outputs were compared with those of the base model to examine whether fine-tuning improved consistency with KG-guided temporal states.

### F. *Qualitative Analysis of KG-Guided Temporal Interpretation*

To analyze model behavior in representative rolling-window cases, we conducted a qualitative comparison using human-guided references. The human-guided reference was constructed primarily from human-provided contextual information and further refined using sensor-derived temporal evidence, including baseline-comparison states and within-window trend states. The resulting reference was manually checked against the corresponding sensor evidence and available daily-context information before being used for model comparison. Therefore, the reference is used to evaluate whether the model interpretation remains consistent with the combined human contextual information and structured temporal evidence. Table III shows examples of the comparison between the base Qwen and the P95-LoRA Qwen under KG-guided interpretation. Each case contains the input rolling-window evidence, baseline states, trend states, model predictions, and the final comparison summary.

TABLE III. QUALITATIVE COMPARISON OF BASE QWEN AND P95-LORA QWEN UNDER FIVE AND SEVEN DAYS ROLLING-WINDOW SETTINGS

| Case | Window Size | Human-Guided Reference | Base Qwen | P95-LoRA Qwen |
|---|---|---|---|---|
| 1 | 5 | Short term fluctuation | Movement fragmentation rising | **Short term fluctuation** |
| 2 | 7 | Short term fluctuation | Movement fragmentation rising | **Short term fluctuation** |

With a 5-day window, the base Qwen model predicted rising movement fragmentation. However, the P95-LoRA Qwen predicted short-term fluctuation, which was closer to the human-guided reference. The input evidence showed several activity changes within the five-day window, but the baseline-related states remained close to the usual pattern. By using KG-based evidence, the fine-tuned model avoided interpreting local short-term changes as a persistent behavioral shift.

With a 7-day window, a similar tendency was observed at the weekly scale. The base Qwen model again predicted rising movement fragmentation, mainly focusing on the increasing workspace-related evidence. In contrast, the P95-LoRA Qwen model predicted short-term fluctuation by considering additional KG evidence, including decreasing entry activity and stable rest time. This suggests that the KG-guided model used a broader set of temporal relations rather than relying on a single increasing feature.

### G. *Holiday Case Study*

To further examine whether the proposed framework can describe lifestyle changes during a special period, we conducted an additional case study using a holiday-related period in the local Home Assistant data. This case study examines whether sensor-derived temporal states can reflect observable lifestyle changes during a period that includes holiday days. A five-day rolling window and a seven-day rolling window were compared to observe how the temporal scale affects the interpretation. The results are summarized in Table IV.

TABLE IV. CASE STUDY COMPARISON DURING A HOLIDAY-RELATED PERIOD

| Case | Window Size | Human-Guided Reference | Base Qwen | P95-LoRA Qwen |
|---|---|---|---|---|
| 1 | 5 | Transition to new pattern | Movement fragmentation rising | **Transition to new pattern** |
| 2 | 7 | Gradually work intensive | Movement fragmentation rising | **Gradually work intensive** |

Based on the seven-day holiday-related case in Table IV, Fig. 5 visualizes the corresponding KG context. The graph shows that behavior features, baseline-comparison evidence, and trend evidence were aggregated into the KG context to support the work-intensive temporal state.

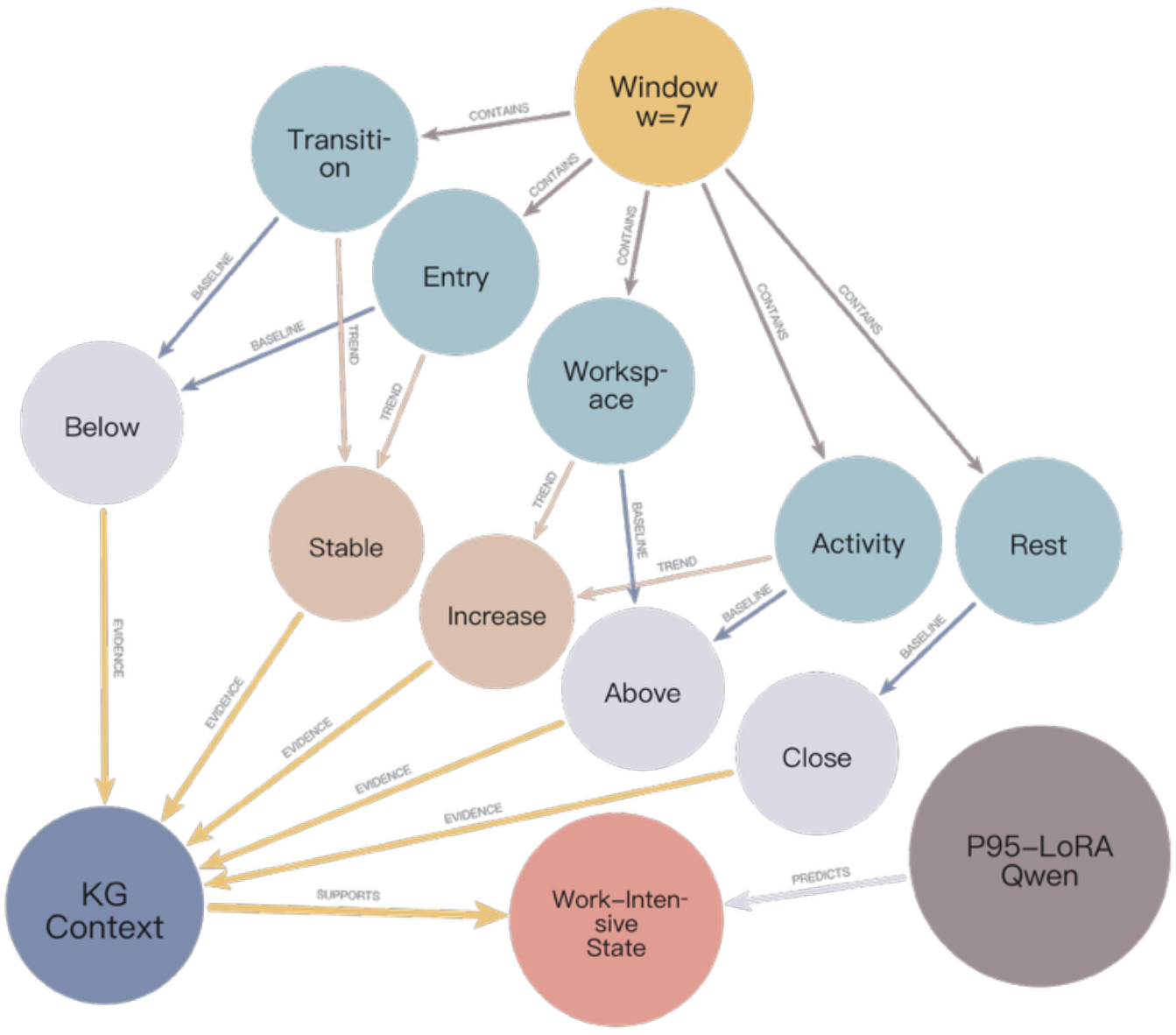

Fig. 5. KG context visualization for the seven-day holiday-related rolling-window case.

In both cases, the base Qwen model predicted a movement-fragmentation pattern, suggesting that it tended to react to local activity or movement-related changes. In contrast, P95-LoRA Qwen produced predictions that were consistent with the human-guided reference in both window settings. In the five-day window, the model identified a transition toward a new behavior pattern, where work-area activity increased while rest time decreased. In the seven-day window, the model predicted a gradual work-intensive pattern, as the longer temporal context suggested a more stable increase in work-related behavior rather than a broader lifestyle transition. These results indicate that the KG-guided model can distinguish different types of lifestyle changes during a holiday-related period without relying on an explicit holiday label.

## V. Discussion

P95 provided a more balanced setting by preserving more behavioral diversity while maintaining relatively stable model performance. The window-size comparison indicates that temporal scale is important for smart-home behavior interpretation. Five-day and seven-day windows produced more stable KG-guided interpretations than shorter windows, suggesting that medium-length windows provide sufficient context to distinguish temporary fluctuations from more persistent behavioral changes.

The base Qwen model tended to focus on local feature changes, such as increased activity or movement-related indicators, and sometimes interpreted them as movement-fragmentation patterns. In contrast, P95-LoRA Qwen used structured evidence such as baseline closeness, trend direction, zone-level behavior, and temporal-state relations. However, there are still some limitations. First, although the local deployment covered 48 days, only the first 22 days were used for local model adaptation, while the subsequent 26 days were used as the held-out local test period. Longer-term evaluation is still needed to examine robustness over broader behavioral and seasonal variations. Second, although the CASAS Aruba dataset was used as an external dataset, its sensor configuration and living environment are different from the local Home Assistant prototype. Therefore, additional smart-home datasets are needed to verify the proposed method's generality. Finally, the current temporal-state reference is primarily based on human-provided contextual information and further refined using sensor-derived temporal evidence. Future work should incorporate more systematic human annotations to further validate the generated interpretations.

## VI. Conclusion

Smart-home behavior monitoring is an important research direction for understanding daily living patterns from residential sensor data. This study aimed to improve the interpretation of temporal behavior changes in smart-home environments by proposing a KG-guided local LLM framework. The method converts low-level Home Assistant sensor histories into daily behavior summaries, constructs rolling-window temporal representations, and uses baseline comparisons and within-window trends to build lightweight KG-based evidence for local LLM interpretation. Experimental results using a local Home Assistant prototype and the CASAS Aruba dataset showed that P95-LoRA Qwen produced more stable and context-aware temporal interpretations than the base Qwen model while generating natural-language explanations grounded in structured temporal evidence. Future work will focus on refining temporal-state decision rules and increasing training examples for ambiguous transition cases to better discriminate similar behavior patterns. Additional testing on smart-home datasets with different room layouts, sensor configurations, and resident behaviors will be conducted to examine the generality of the framework beyond the CASAS Aruba dataset and the local prototype. Further evaluation of lightweight language models and different LoRA configurations will also be performed to determine whether the KG-guided interpretation method generalizes beyond Qwen.